\documentclass{article}

\usepackage{microtype}
\usepackage{amsmath}
\usepackage{amsfonts}
\usepackage{graphicx}
\usepackage{subfigure}
\usepackage{booktabs} 
\usepackage{csquotes}
\usepackage{siunitx}
\usepackage{dcolumn}
\usepackage{xr}
\usepackage{hyperref}

\usepackage[accepted]{icml2018}

\include{macros}

\icmltitlerunning{Erratum Concerning the Obfuscated Gradients Attack on Stochastic Activation Pruning}

\begin{document}

\twocolumn[
  \icmltitle{Erratum Concerning the Obfuscated Gradients Attack \\ on Stochastic Activation Pruning}

\icmlsetsymbol{equal}{*}

\begin{icmlauthorlist}
\icmlauthor{Guneet S. Dhillon}{equal,a}
\icmlauthor{Nicholas Carlini}{equal,b}
\end{icmlauthorlist}

\icmlaffiliation{a}{Amazon Web Services (this work is not related to the author's employment)}
\icmlaffiliation{b}{Google Brain}



\vskip 0.25in
]



\printAffiliationsAndNotice{\icmlEqualContribution} 

\begin{abstract}
  Stochastic Activation Pruning (SAP) \citep{dhillon2018stochastic} is a defense to adversarial examples
  that was attacked and found to be broken by the ``Obfuscated Gradients'' paper \citep{athalye2018obfuscated}.
  We discover a flaw in the re-implementation
  that artificially weakens SAP.
  When SAP is applied properly, the proposed attack is not effective.
  However, we show that a new use of the BPDA attack technique
  can still reduce the accuracy of SAP to $0.1\%$.
\end{abstract}

\section{Introduction}

Stochastic Activation Pruning (SAP) \citep{dhillon2018stochastic} is a proposed defense
to adversarial examples.
In their work, \citet{athalye2018obfuscated} perform an analysis of 
SAP and determine that it offers no robustness improvement on top of a baseline model.
We discover a flaw in the re-implementation made by the authors that 
artificially weakens SAP.
A different attack technique is necessary to break the correctly-implemented version of SAP.

\section{Background}

We assume familiarity with neural networks,
methods to generate adversarial examples,
Stochastic Activation Pruning,
and 
the Backwards Pass Differentiable Approximation.

\textbf{Notation.}
For a trained neural network $f(\cdot)$ evaluated on some
input $x$, an adversarial example $x'$ is constructed by performing gradient
ascent in the input-space to maximize the loss function $\ell(f(x'), y)$
(cross-entropy loss in this case).
This is done with the constraint that the distance between $x$ and $x'$ (the infinity norm is commonly used) is small.

\textbf{Stochastic Activation Pruning (SAP)} \citep{dhillon2018stochastic}
introduces randomness into the evaluation of a pre-trained neural network
by stochastically dropping out neurons and setting their values to zero.
Neurons are retained with
probabilities proportional to their absolute value.

Let $f = f_d \circ f_{d-1} \circ \dots \circ f_1$ denote a $d$-layer neural network.
We define $h^i \in \mathbb{R}^{m_i}$ to be the activations which result after evaluating layer $f_i$ (with the non-linearity).
We index each activation as $h^i_j$.

While performing the forward pass, SAP defines a multinomial probability distribution
\[ p^i_j = |h^i_j| \cdot \bigg(\sum\limits_{k=1}^{m_i} |h^i_{k}|\bigg)^{-1}, \]
where $p^i_j$ is the probability of retaining $h^i_j$. $r_i$ neurons are randomly sampled with replacement
according to this probability distribution.
The probability that $h^i_j$ is retained is
$q^i_j = {1 - (1 - p^i_j)^{r_i}}$.
To ensure that the total ``mass'' propagating forward is preserved,
SAP divides each node by the probability of retaining it (similar to dropout), so that
\[ \hat{h}^i_j = \begin{cases}
  {h^i_j \over q^i_j} & \text{if sampled} \\
  0 & \text{otherwise}. \\
\end{cases}
\]
This process is repeated for every non-linear layer.

The choice of $r_i$ should be
large enough that not too many neurons are dropped (otherwise SAP would not be accurate on clean data), but not so
large that all neurons are retained (otherwise SAP would do
nothing).
The authors suggest setting $r_i$ to be equal to the
width of the layer, i.e. $m_i$.

\textbf{Backwards Pass Differentiable Approximation (BPDA)}
\citep{athalye2018obfuscated} is an attack strategy that alters
the computation of the gradient of $f$ with respect
to the input $x$, i.e. $\nabla_x f(x)$.
The forward pass is computed on the function $f$, but 
the backward pass is computed on a different function $g \approx f$ such that the
resulting gradient is neither the gradient of $f$ nor the gradient of $g$.

Specifically, let $f^{i}$ be a non-differentiable
layer of a neural network.
To approximate $\nabla_x f(x)$,
construct an approximation $g^i\approx f^i(x)$ of this layer.
Then, approximate $\nabla_x f(x)$ by performing the forward
pass through $f(\cdot)$ (in particular, $f^i(x)$),
but on the backward pass, replace $f^i(x)$ with $g^i(x)$.
In general when multiple layers are non-differentiable we select one
$g^i$ per layer, and replace all of them in the backward pass.
As long as the two functions are similar, the slightly inaccurate
gradients still prove useful in constructing adversarial examples.

\section{The Error of ``Obfuscated Gradients ...''}

The SAP paper
explicitly states ``the output of stochastic models are computed as
an average over multiple forward passes'' \citep{dhillon2018stochastic}.
When re-implementing the SAP defense, the authors of \citet{athalye2018obfuscated} did not include this step\footnote{The author of this erratum, Nicholas Carlini, wrote the SAP re-implementation and is solely responsible for the error.}.
As a result, in order to maintain the clean accuracy of approximately $83\%$ as reported
in \citet{dhillon2018stochastic}, the value of $r_i$ had to be set to $2 \times m_i$,
which is much larger than the prescribed value.
Fixing this error in the implementation is simple: setting the value of $r_i$ to $m_i$, and evaluating each test example by averaging the outputs over $100$ forward passes.

When this error is corrected, the attack described in \citet{athalye2018obfuscated} is no
longer effective; the accuracy of SAP remains as is claimed in the paper.

Importantly, this error would not have been discovered if not for the
fact that both papers \citep{dhillon2018stochastic,athalye2018obfuscated} released source code.
We firmly believe that releasing source code is the only way to promote
correct and reproducible research, especially in the domain of adversarial
machine learning where -- as is the case here -- setting a single hyper-parameter
to the incorrect value can have dramatic consequences.

There was a second difference that did not change the results.
Instead of sampling exactly $r_i$ neurons per layer from a multinomial distribution as in \citet{dhillon2018stochastic},
\citet{athalye2018obfuscated} used a per-neuron binomial distribution.
This approximation is more efficient in high-dimensional spaces while remaining close in performance.
When we attack a model that uses the latter approach,
and evaluate using the former, the attack success
rate remains unchanged.

\section{Repairing the SAP Attack}

When we run the attack code on 
the correctly-implemented version of SAP, 
it fails to find
an adversarial example in most cases; even with over $10,000$ iterations of gradient
ascent, the targeted attack success rate remains below $50\%$ on CIFAR-10 at a distortion bound of  $\varepsilon=0.031$.

We therefore began to test for other signs of gradient masking as recommended
by \citet{athalye2018obfuscated}.
We ran a \emph{transfer attack} where we generated adversarial examples
on the undefended model and then evaluate these adversarial examples with SAP.
The targeted attack success rate is $70\%$ on these.
Part of the reason why this attack is more successful is due to gradient masking.
Intuitively this makes sense as \emph{SAP
introduces stochasticity on top of a pre-trained model},
behaving similar to the pre-trained model
while making the attack optimization difficult.

Given that gradients computed on the undefended model effectively fool the defended model, we decided to try and apply the
BPDA\footnote{BPDA in general should
not be treated as a magic black-box that resolves all optimization difficulties.
Since BPDA applies the incorrect gradient, in many cases unless applied very carefully,
attacks perform \emph{worse} with BPDA than without.
In this particular case we found it was helpful.}
strategy on SAP.
By doing this, we query the actual defended model, while only taking gradients
with respect to the original model.
The concrete instantiation of this attack removes the neuron-dropping completely from the backward pass and just computes the gradients on the
vanilla neural network $f$, without any SAP components; the forward pass retains the
dropped neurons.
As mentioned earlier, we apply per-neuron binomial sampling for efficiency, but 
test on the correct multinomial distribution.

We then evaluate the accuracy of SAP on CIFAR-10
with a distortion bound of $\varepsilon=0.031$.
This modified attack is sufficient to reduce the accuracy of SAP to $0.1\%$ ($\pm 0.05\%$) evaluated over the test set.

\section{Conclusion}

We discover a flaw in the evaluation of \citet{athalye2018obfuscated} with regard
to the implementation of Stochastic Activation Pruning \citep{dhillon2018stochastic}. When corrected, the
original attack is no longer effective. However, we slightly adapt the attack
to make use of BPDA and reduce the effectiveness of SAP to $0.1\%$ 
at $\varepsilon=0.031$.

Papers which re-implement defenses must be extremely careful when reproducing
prior work to ensure that any replications are exactly as described as in the
original paper. In this case, the error should have been discovered when the
replicated neural network required a different hyper-parameter than described in the
original paper.
Fortunately, the reason this discrepancy was discovered at all was
that both papers \emph{did} release code (by publication time).

\bibliography{paper}
\bibliographystyle{icml2018}

\end{document}